\documentclass{uai2023} 

\usepackage[american]{babel}

\usepackage{natbib} 
\usepackage{mathtools} 
\usepackage{booktabs} 
\usepackage{tikz} 
\usepackage{microtype}
\usepackage{graphicx}
\usepackage{subfigure}
\usepackage{wrapfig}
\usepackage{hyperref}
\usepackage{algorithmic}
\usepackage{algorithm}

\usepackage{amsmath}
\usepackage{amssymb}
\usepackage{mathtools}
\usepackage{amsthm}

\usepackage[capitalize,noabbrev]{cleveref}

\DeclareMathOperator*{\argmin}{arg\,min}

\renewcommand{\algorithmicrequire}{\textbf{Input:}}
\renewcommand{\algorithmicensure}{\textbf{Output:}}

\newcommand{\ind}[2]{#1^{(#2)}}

\newcommand{\R}{\mathbb{R}}

\newcommand{\E}{\mathbb{E}}

\newcommand{\ith}[1]{#1^\mathrm{th}}

\newcommand{\irange}[2]{\{#1, \ldots, #2\}}
\newcommand{\xind}[1]{x^{(#1)}}

\newcommand{\refsec}[1]{\S\,\ref{#1}}
\newcommand{\refalg}[1]{Alg.~\ref{#1}}

\newcommand{\reffig}[1]{Fig.~\ref{#1}}

\usepackage{letltxmacro}
\LetLtxMacro{\originaleqref}{\eqref}
\renewcommand{\eqref}{eq.~\originaleqref}

\theoremstyle{plain}

\theoremstyle{definition}

\theoremstyle{remark}

\usepackage[disable,textsize=tiny]{todonotes}
\usepackage{scalerel}

\title{Acquisition Conditioned Oracle for Nongreedy Active Feature Acquisition}

\author[1]{{Michael Valancius}{}}
\author[2]{Max Lennon}
\author[2]{Junier B. Oliva}
\affil[1]{%
    Dept. of Biostatistics\\
    University of North Carolina\\
    Chapel Hill, North Carolina, USA
}
\affil[2]{%
    Dept. of Computer Science\\
    University of North Carolina\\
    Chapel Hill, North Carolina, USA
}

\begin{document}
\maketitle

\begin{abstract}
We develop novel methodology for active feature acquisition (AFA), the study of how to sequentially acquire a dynamic (on a per instance basis) subset of features that minimizes acquisition costs whilst still yielding accurate predictions.  
The AFA framework can be useful in a myriad of domains, including health care applications where the cost of acquiring additional features for a patient (in terms of time, money, risk, etc.) can be weighed against the expected improvement to diagnostic performance.
Previous approaches for AFA have employed either: deep learning RL techniques, which have difficulty training policies in the AFA MDP due to sparse rewards and a complicated action space; deep learning surrogate generative models, which require modeling complicated multidimensional conditional distributions; or greedy policies, which fail to account for how joint feature acquisitions can be informative together for better predictions.
In this work we show that we can bypass many of these challenges with a novel, nonparametric oracle based approach, which we coin the acquisition conditioned oracle (ACO).
Extensive experiments show the superiority of the ACO to state-of-the-art AFA methods when acquiring features for both predictions and general decision-making.
\end{abstract}

\section{Introduction}

An overwhelming bulk of efforts in machine learning are devoted to making predictions when given a fully observed feature vector, $x \in \R^d$. Although this paradigm has produced impressive models, it largely ignores that the collection of features comes at a cost, since it serves up all the possible features up front (requiring a potentially expensive complete collection  of features for all instances). 
In contrast, real-world scenarios frequently involve making judgements with some features or information missing.
In a conceptual leap beyond conventional imputation and partially observed methodology, here we note that many situations allow for the dynamic collection of information on an instance at inference time. 
That is, often one may query for additional information (features) that are currently missing in an instance to make a better prediction.
This is especially so in an automated future, where autonomous agents can routinely interact with an environment to obtain more information as they make decisions.
In this work we develop novel, more performant methodology for \emph{active feature acquisition} (AFA), 
the study of how to sequentially acquire a dynamic (on a per instance basis) subset of features that minimizes acquisition costs whilst still yielding accurate predictions.


Consider the following illustrative applications where AFA is useful. 
First, we consider an automated survey system to perform psychological assessments. A traditional ML approach would require collecting responses on an exhaustive list of survey questions, and only after all responses (features) are collected would one make a prediction of the correct assessment. 
However, administering a long survey is slow, may lead to user fatigue \citep{early2016cost}, and may even decrease the accuracy of responses \citep{early2016test}.
Instead, an AFA approach would \emph{sequentially} decide what next question (if any) to prompt the user with to help it make its assessment, ultimately leading to a prediction with a succinct, personalized subset of features per instance.
Note that in contrast to traditional feature selection, which would always select the same subset of questions, an AFA approach will determine a custom subset of questions (of potentially different cardinalities) to ask on a per case basis.
This is because, in AFA, the next feature (answer to a question) to acquire can depend on the values of previous acquisitions. For example, if an instance answers ``no'' to ``Are you sleeping well?'' the next acquisition may be for ``Did you go outside today?''; had they answered ``yes'' instead, the next acquisition could be for ``Are you eating well?''. 
Similar applications are possible for educational assessments and automated trouble shooting systems. In cyberphysical systems, AFA approaches can be used to determine what sensors are necessary (and when they are necessary), potentially reducing latency and extending battery life.

\vspace{0pt}
\begin{figure}
    \centering
    \includegraphics[trim={20mm 0mm 0mm 0mm},clip,width=0.23\textwidth]{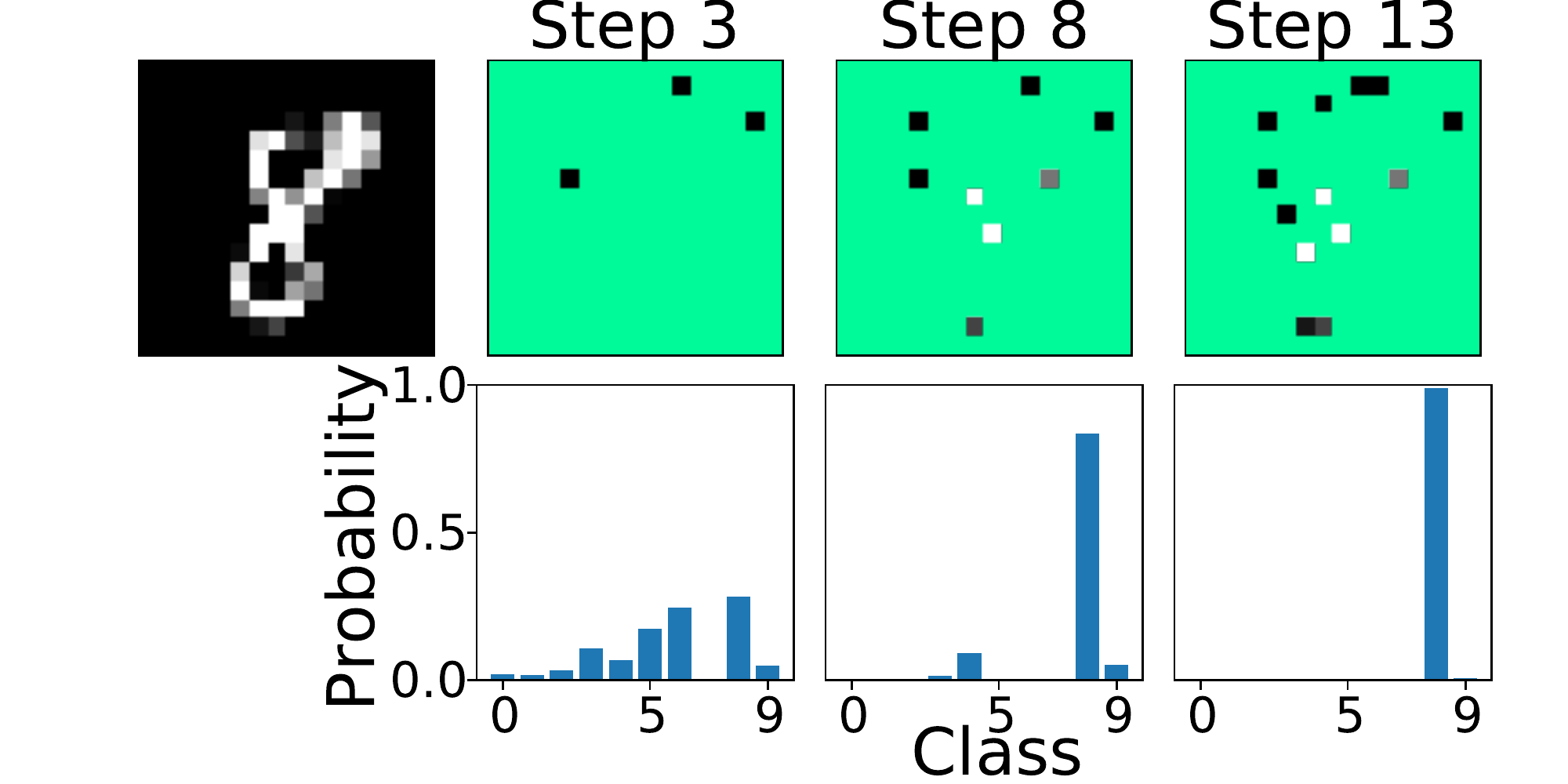}
    \includegraphics[trim={0mm 0mm 0mm 0mm},clip,width=0.165\textwidth]{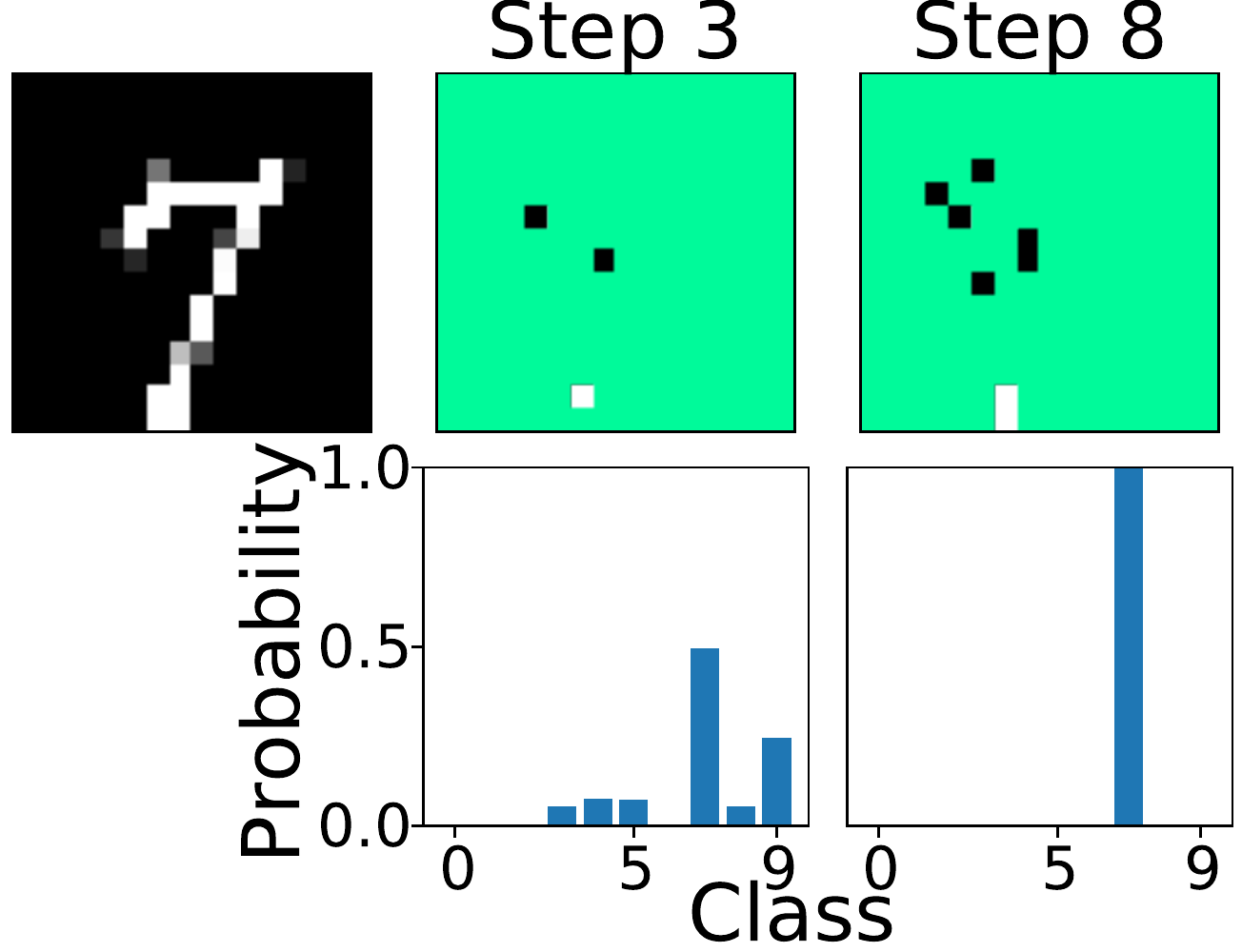}
    \vspace{-6pt}
    \caption{Acquisition process (top) and the prediction probabilities (bottom) using ACO agent on 2 MNIST digits (\refsec{sec:real}). 
    Features are acquired sequentially and vary per instance. 
    }
    \vspace{0pt}
    \label{fig:afa_acq}
\end{figure}
In this work, we introduce the Acquisition Conditioned Oracle (ACO), a nonparametric 
approach to perform active feature acquisition. Through a careful design, ACO presents a novel yet simple approach that has many advantages over predecessors. Namely, ACO is a: non-greedy, deployable oracle that can be used to train a parametric policy, all without the challenging task of reinforcement learning \citep{minsky1961steps,sutton1988learning, shim2018joint} or generative surrogate model training \citep{gu2016continuous, ma2018eddi,li2021active}, whilst achieving better empirical results. In addition to active feature acquisitions for making predictions, we generalize the ACO with novel techniques for active feature acquisition when making \emph{decisions} to maximize expected outcomes.

Our contributions are as follows:
1) we develop the ACO, a novel, non-greedy oracle that is based on conditional dependencies in  feature and labels;
2) we develop a data-driven approximation to the ACO based on simple nonparametric techniques;
3) we develop a simple approach to train a parametric policy for AFA using the ACO;
4) we extend the ACO to decision-making tasks where the policy attempts to maximize expected outcomes based on actively acquired features;
5) we show the empirical effectiveness of our approach 
compared to more complicated state-of-the-art (SOTA) deep learning based AFA models.

\section{Background}

We now expound on the formal definition of the AFA problem. Throughout, we consider underlying instances $x\in\R^d$ and corresponding labels $y$. We denote the $\ith{i}$ feature as $x_i \in \R$, and a subset of feature values $o \subseteq \{1, \ldots, d\}$ as $x_o \in \R^{|o|}$. 
At a high level, the AFA problem is concerned with a sequential, dynamic acquisition of salient features, such that one can make a successful prediction based only on the acquired feature values.
As several previous works have noted \citep{zubek2002pruning, ruckstiess2011sequential, shim2018joint}, AFA can be neatly encapsulated as the following MDP:
states $s=(x_o, o)$ are comprised of the currently observed features $o$, and the respective values $x_o$; actions $a \in \left(\{1, \ldots, d\} \setminus o \right) \cup \{\phi\}$ indicate whether to acquire a new feature value, $a \in \{1, \ldots, d\} \setminus o $, or to terminate with a prediction, $a = \phi$; when making a prediction ($a=\phi$) rewards are based on a supervised loss $-\ell(\hat{y}(x_o, o), y)$ of a prediction based on observed features $\hat{y}(x_o, o)$ and the ground truth label $y$, otherwise the reward is a negative cost of acquiring another feature onto $o$, $-c(a, o)$ (commonly at a constant cost $c(a, o) = \alpha$); lastly, for non-terminal actions, the state transitions $(x_o, o) \rightarrow (x_{o \cup \{a\}}, o \cup \{a\})$.
Note that predicting $y$ is done via an estimator $\hat{y}(x_o, o)$ (abbreviated as $\hat{y}(x_o)$) that is able to take in various (arbitrary) subsets of features to infer the label. In practice, this estimator is often trained ahead of time (and possibly fine-tuned) using a neural network with a masking scheme \citep{li2021active} or 
set-embedding of observed features \citep{ma2018eddi}. 

Although an agent policy $\pi(x_o, o)$ for the aforementioned MDP may be trained directly using reinforcement learning techniques \citep{ruckstiess2011sequential, shim2018joint}, the large action space and sparse reward structure make this a difficult RL task. To assuage these issues, recent work \citep{li2021active} has employed a learned surrogate generative model and model-based RL techniques to improve the learned agent policies. Other approaches have bypassed training a general policy and instead utilize greedy policies to select features. For instance, \cite{ma2018eddi, gong2019icebreaker} use a generative approach to estimate the expected utility (e.g., mutual information) of any one acquisition, employing a greedy strategy to maximize utility; similarly \citet{zannone2019odin} also consider information gain for AFA with limited dataset sizes. Most related to our work, \cite{he2012imitation, he2016active} utilize a greedy ``oracle'' and imitation learning techniques to teach a supervised policy to select features. In \refsec{sec:meth} we develop an alternative, non-greedy oracle, which may also be used to supervised a parametric policy.

\subsection{Related Works}
Before developing our methodology, we briefly compare and contrast AFA to other related directions in ML.

\textbf{Feature Selection}\quad
Feature selection \citep{chandrashekar2014survey, miao2016survey, li2017feature, cai2018feature}, ascertains a \emph{constant} subset of features that are predictive.
Feature selection methods choose a \emph{fixed} subset of features $s \subseteq \irange{1}{d}$, and always predict $y$ using this same subset of feature values, $x_s$.
In contrast, AFA considers a \emph{dynamic} subset of features that is sequentially chosen and personalized on an instance-by-instance basis (\reffig{fig:afa_acq}) to increase useful information. 
Feature selection can be seen as a special case of AFA (perhaps a stubborn one) that selects the same next features to observe regardless of the previous feature values it has encountered.
(It is worth noting that AFA may be applied after an initial feature selection preprocessing step to reduce the search space.)
In general, a dynamic strategy will be helpful whenever covariates are indicative of other features that are dependant with the output, which is often the case in real-world examples
\footnote{A simple case that illustrates the dominance of an active acquisition strategy is as follows. Suppose that there are $d-1$ independent features $x_i$ for $i\in \{1, \ldots, d-1\}$, and one ``guiding'' feature, $x_d$, which decides what independent feature determines the output $y$. E.g., if $x_d$ is in some range then $y$ is a function of only $x_1$, if $x_d$ is in another range then $y$ is a function of $x_2$, etc. If all independent features are used with equal probability, then a typical feature selection algorithm will have to select all $d$ to obtain a high accuracy (since it must use a fixed subset). In contrast, by dynamically selecting features, AFA can observe only two features: $x_d$, and the corresponding independent feature.}.

\textbf{Active Learning}\quad
Active learning (e.g.,~\cite{fu2013survey, konyushkova2017learning, yoo2019learning, mackay1992information,houlsby2011bayesian}) is a related task in ML to gather more information when a learner can query 
an oracle for the \emph{true label}, $y$, of
a \emph{complete} feature vector $x \in \R^d$ to \emph{build a better estimator} $\hat{f}$. 
Active learning considers queries to an \emph{expert} for the correct \emph{output} label to a \emph{complete} set of features in order to construct a training instance to \emph{build a better classifier}.
Instead, AFA considers queries to \emph{the environment} for the \emph{feature value} corresponding to an unobserved feature dimension, $i$, in order to provide a \emph{better prediction on the current instance}.
Thus, while the active learning paradigm queries an \emph{expert} during \emph{training} to build a classifier with complete features, the AFA paradigm queries the \emph{environment} at \emph{evaluation} to obtain unobserved features of a current instance to help its current assessment.

\section{Methods}
\label{sec:meth}

We develop our methodology by first designing an oracle that operates over the same information as the AFA policy (an important restriction that has not been previously respected); after, we develop a data-driven approximation to the oracle, and show how to learn a parametric student policy to emulate it. Lastly, we extend our approach to also consider the active acquisition of features when making decisions that maximize expected general outcomes.

\subsection{Cheating Oracle}

We begin by considering previous retrospective oracle approaches \citep{he2012imitation, he2016active,madasu2022learning}, which greedily determine what next feature would ideally be acquired for a particular instance $x$, with corresponding label $y$, given an observation set $o$ of already acquired features. 
That is, such a retrospective approach will look for the next feature, $i$, that, when added to the already acquired features, $o$, results in the best improvement of a loss $\ell$ (e.g., MSE or cross-entropy) using an estimator $\hat{y}$:
\begin{equation}
    i = \argmin_{j \in \{1, \ldots, d\} \setminus o} \ell\left( \hat{y}(x_{o \cup \{j\}}), y \right). \label{eq:greedy_search}
\end{equation}
Although this approach has been previously described as an ``oracle'' (e.g., the ``forward-selection oracle,'' \cite{he2012imitation}), such a designation is somewhat misleading, as this search operates over information that is not available to the agent.
Note that to select the next action,  \eqref{eq:greedy_search} will try the classifier (or regressor) $\hat{y}$ on the instance $x$ with the instance's values $x_{o \cup \{j\}}, \ \forall j \in \{1, \ldots, d\} \setminus o$ (for yet to be acquired features), and compare the predictions to the instance's true label $y$; this requires complete knowledge about $x$ (as well as its true label $y$). Thus, in actuality, the resulting action from \eqref{eq:greedy_search} depends on inputs $x$ (the entire feature vector), $y$ (the respective true label), and $o$ (the already acquired feature indices): $\pi_\mathrm{cheat}(x, y, o) = \argmin_{j \in \{1, \ldots, d\} \setminus o} \ell\left( \hat{y}(x_{o \cup \{j\}}), y \right)$.
As an oracle is meant to provide an optimum action for the same state space, we see that \eqref{eq:greedy_search} ``cheats,''  since it operates over more information than the AFA agent has available, $\pi(x_o, o)$.

Although the cheating oracle offers an intuitive appeal and variants of it have been used as a reference teacher policy in AFA \citep{he2012imitation, he2016active,madasu2022learning}, here we develop an alternative oracle approach that:
\begin{enumerate}[topsep=0mm,itemsep=1mm,partopsep=0mm,parsep=0mm,labelsep=2mm,labelwidth=\itemindent,leftmargin=2mm,align=left]
    \item \textbf{Is Non-greedy}: Previous retrospective approaches have looked for one single acquisition to make at a time. 
    However, once the agent decides to make a prediction, it uses its acquired features \emph{jointly} to estimate the output. 
    Greedily acquiring the next best feature ignores jointly informative groups of multiple features that may be acquired.
    Thus, a better strategy (one that we will employ) considers acquisitions \emph{collectively} for the final prediction.
    \item \textbf{Respects Sequential Acquisition}:  The cheating oracle \newline operates over more information than is accessible to an AFA agent, since it utilizes not-yet-acquired features (and the true label) while evaluating feature subsets
    and implicitly uses this information to select the next
    feature.
    In contrast, the AFA agent $\pi(x_o, o)$ only has access to the already acquired feature values $x_o$ when selecting the next action. 
    That is, the agent cannot make a decision based on information it has not acquired. 
    In order to more directly provide guidance on the correct acquisition action to take when provided with \emph{partially observed} information on an instance, we shall define a non-cheating oracle that operates only over acquired feature values, which should ultimately provide more accurate targets to a learned student policy.
    \item \textbf{Is Deployable}: 
    In principle, an oracle should be deployable in one's environment, as it should be able to yield the correct action given \emph{the same information} (the same states) as the agent. As a consequence of utilizing information that is not available during a roll-out
    , the cheating oracle is not deployable at inference time, making it difficult to judge its performance.
    Instead, our approach shall yield a deployable oracle, making it possible to judge the performance of the oracle at test time and disentangling
    any degradation in performance due to imitation-learning.
\end{enumerate}

\subsection{Acquisition Conditioned Oracle}
Here we now define our non-cheating, deployable oracle, which we coin the \emph{acquisition conditioned oracle} (ACO). To begin, we define the ACO in terms of a full distributional knowledge over features and labels $p(x,y)$ (including all marginal and conditional distributions), and later describe approximations based on training data. First, we generalize the greedy search in \eqref{eq:greedy_search} to a non-greedy approach using a weighted (by $\alpha > 0$) cost $c$ for subsets:
\begin{equation}
    u(x, y, o) = \argmin_{v \subseteq \{1, \ldots, d\} \setminus o} \ell\left( \hat{y}(x_{o \cup v}), y \right) + \alpha c(o \cup v). \label{eq:nongreedy_search}
\end{equation}
The (still ``cheating'') \eqref{eq:nongreedy_search} now considers additional subsets of features $u \subseteq \{1, \ldots, d\} \setminus o$ atop of $o$ to best reduce the loss along with the cost of the additional features $c(o \cup u)$. 
The search-space is now much larger than the greedy search \eqref{eq:greedy_search}; below we describe effective strategies to navigate the large search.
In addition, \eqref{eq:nongreedy_search} still operates with unacquired features of $x$ in $\{1, \ldots, d\} \setminus o$, and thus is still not an apples-to-apples (in the state-space) oracle to  the AFA policy. 
We propose an intuitive, straightforward, and ultimately effective, \emph{novel} approach to avoid utilizing unacquired features:
use the conditional distribution (conditioned on acquired features) over labels and unacquired features,
\begin{equation}
    \resizebox{.9\hsize}{!}{$u(x_o, o) = \argmin\limits_{v \subseteq \{1, \ldots, d\} \setminus o} \E_{\mathbf{y}, \mathbf{x}_v | x_o} \left[ \ell\left( \hat{y}(x_{o}, \mathbf{x}_v), \mathbf{y} \right) \right] + \alpha c(o \cup v)$}. \label{eq:aco_search}
\end{equation}
At a high-level, this approach imagines \emph{likely scenarios} of the unacquired feature values and labels (based on the acquired features and their values), and determines additional subsets of features to acquire based on these imagined values. Whilst a natural choice (in retrospect) for how to determine useful acquisitions without ``cheating,'' we emphasize the novelty of this oracle.
\todo{could have detail on $\hat{y}(x_{o}, \mathbf{x}_v)$}
Whenever the ACO minimization \originaleqref{eq:aco_search} returns $u(x_o, o) = \varnothing$, it is clear that there are no further acquisitions that are worth the cost.
Below we discuss alternatives for when $u(x_o, o) \neq \varnothing$, which indicates that (based on the acquired information) we would like to acquire all the features in $u(x_o, o)$ jointly to make a prediction. As the typical AFA MDP acquires one feature at a time, a simple approach is to uniformly select one acquisition from the subset $u(x_o, o)$. That is, we may define the ACO's policy (\refalg{alg:aco_alg}) as (with notational abuse for clarity):
\begin{align}
    &\pi_{\mathrm{ACO}}(x_o, o) =  \nonumber \\
    & \resizebox{.9\hsize}{!}{$\mathbb{I}\{u(x_o, o) = \varnothing\} \hat{y}(x_{o})
    + \mathbb{I}\{u(x_o, o) \neq \varnothing\} \mathrm{Unif}[u(x_o, o)]$}.
    \label{eq:aco_policy}
\end{align}
As the loss is in expectation over the unobserved label and features \originaleqref{eq:aco_policy}, the ACO avoids ``cheating'' and plugging in unacquired values not available during inference as \eqref{eq:greedy_search}.


\begin{algorithm}[tb]
   \caption{Acquisition Conditioned Oracle}
   \label{alg:aco_alg}
\begin{algorithmic}
   \STATE {\bfseries Input:} Initial observed features $o$ (possibly $o=\varnothing)$, instance values $x_o$, joint distribution $p(x,y)$, estimator $\hat{y}$
   \STATE Initialize \texttt{predict} $\coloneqq$ \texttt{false}
   \WHILE{$|o| < d$ {\bfseries and  not} \texttt{predict}}
   \STATE $u(x_o, o) \coloneqq$ \resizebox{.7\hsize}{!}{$\argmin\limits_{v \subset \{1, \ldots, d\} \setminus o} \E_{\mathbf{y}, \mathbf{x}_v | x_o} \left[ \ell\left( \hat{y}(x_{o}, \mathbf{x}_v), \mathbf{y} \right) \right] + \alpha c(o \cup v)$}
   \IF{$u(x_o, o) = \varnothing$}
   \STATE \texttt{predict} = \texttt{true}
   \ELSE 
   \STATE $j \sim \mathrm{Unif}[u(x_o, o)]$, $o \coloneqq o \cup \{j\}$
   \ENDIF
   \ENDWHILE
   \STATE {\bfseries Return} Prediction $\hat{y}(x_o)$
\end{algorithmic}
\end{algorithm}

\subsection{Approximate ACO}

There are two main limitations that make the ACO minimization \originaleqref{eq:aco_search} infeasible in practice: 1) the ground truth data distribution is unknown; and 2) the search space over subsets $v \subseteq \{1, \ldots, d\} \setminus o$ is large for most dimensionalities.
While both problems have many potential solutions, we show (see empirical results, \refsec{sec:exps}) that simple, straightforward strategies are performant. Throughout, we assume that we have a training dataset $\mathcal{D} = \{(\xind{i}, \ind{y}{i})\}_{i=1}^{n}$ of $n$ input/output tuples as is common in prior AFA approaches.

First, we note that in practice the data distribution, $p(x,y)$, is unknown and therefore cannot be used in the ACO minimization \originaleqref{eq:aco_search} and resulting policy \originaleqref{eq:aco_policy}. 
Furthermore, since the data distribution must be conditioned as $p(y, {x}_v | x_o)$, for $v \subseteq \{1, \ldots, d\} \setminus o$ (e.g., for the expectation in \eqref{eq:aco_search}), we must be able to condition on \emph{arbitrary} subsets of features $o$, which is typically infeasible for many performant generative models \citep{kingma2016improving, dinh2, oliva2018transformation}. Advances in \emph{arbitrary conditional} models \citep{ivanov2018variational,belghazi2019learning,li2020acflow,molina2019spflow,strauss2021arbitrary}, which are able to approximate conditional distributions $p(x_u \mid x_o)$ for arbitrary subsets $u,o$, and recent work that leverages these \emph{arbitrary conditional} model to provide a surrogate model-based RL approach for AFA \citep{li2021active}, suggest that one can learn distribution approximators to draw realistic samples over $\mathbf{y}, \mathbf{x}_v | x_o$, for $v \subseteq \{1, \ldots, d\} \setminus o$. However, this approach requires learning an expensive generative model, which can be challenging due to computation, hyperparameter-optimization, and sample-complexity. We propose to bypass a complicated generative approach and instead sample labels and unacquired features through neighbors, i.e. 
\begin{align}
    \resizebox{.9\hsize}{!}{$\E_{\mathbf{y}, \mathbf{x}_v | x_o} \left[ \ell\left( \hat{y}(x_{o}, \mathbf{x}_v), \mathbf{y} \right) \right] \approx \tfrac{1}{k} \sum\limits_{\scaleto{i \in N_k(x_o)}{5pt}} \ell\left( \hat{y}(x_{o}, \xind{i}_v), \ind{y}{i} \right)$},
\end{align}
where $N_k(x_o)$ is the set of $k$ nearest neighbor indices in $\mathcal{D} = \{(\xind{i}, \ind{y}{i})\}_{i=1}^{n}$, to $x_o$ (i.e., comparing instances only using features $o\subseteq \{1, \ldots, d\}$ to values $x_o$ according to a distance function $d(x_{o}, \xind{j}_{o}) \mapsto \mathbb{R}_+$). In practice, as we typically only consider acquiring small subsets $o$, this approach provides an effective way to generate useful samples\footnote{Note that special care must be taken to avoid returning an instance as its own neighbor when $x_o$ is from the training data.}. A special case arises when $o = \varnothing$ (when starting the acquisition with no features selected),
where all dataset instances are valid neighbors in $N_k(\varnothing)$. 
In this case, the first feature to acquire can be drawn from some fixed distribution $j \sim \pi_0$ based on 
\resizebox{.7\hsize}{!}{$\argmin_{v \subseteq \{1, \ldots, d\} \setminus o} \tfrac{1}{n} \sum\limits_{i=1}^n \ell\left( \hat{y}(\xind{i}), \ind{y}{i} \right) + \alpha c(o \cup v)$}, which needs to be computed only once; alternatively the first feature to acquire atop of an empty set can be set as $j \coloneqq j_0$ where $j_0$ is a hyperparameter that can be validated).

The proposed ACO policy considers all possible additional subsets of features to append to a current set of features $o$ \eqref{eq:aco_search}. This results in a number of possible subsets that grows exponentially in the dimensionality, $d$. When the number of features, $d$, is modestly sized, it is feasible to search across each subset, however this quickly becomes infeasible for larger dimensionalities. We note that minimizing over $ v \subseteq \{1, \ldots, d\} \setminus o$ can be posed as a discrete optimization problem (as it is equivalent to searching over binary membership indicator vectors); hence, one may deploy a myriad of existing discrete optimization approaches \citep{parker2014discrete} including relaxations \citep{pardalos1996continuous}, and genetic algorithms \citep{rajeev1992discrete}. 
Instead, we propose an efficient, simple approach which randomly subsamples potential subsets to search over, $\mathcal{O} \subseteq \{v | v \subseteq \{1, \ldots, d\} \setminus o \}$. Although this is no longer guaranteed to return the optimal subset, we found that in practice a cardinality of $|\mathcal{O}|$ in the few thousands typically performed well (see \refsec{sec:exps}), indicating that slightly suboptimal subsets were still effective.

Lastly, a subtlety arises in developing an AFA policy from the ACO minimization \originaleqref{eq:aco_search} since AFA policies typically acquire one feature at a time. The ACO minimization returns an additional subset of multiple features, $u(x_o, o)$ that are likely to be \emph{jointly} informative at a relatively low cost given the present information $x_o$. As the oracle suggests that ending up acquiring the features $u(x_o, o)$ is beneficial, a sensible heuristic is to choose to acquire any one of the features in $u(x_o, o)$ uniformly at random \originaleqref{eq:aco_policy}. Note that after choosing to acquire a feature $j \in u(x_o, o)$, and observing the value $x_j$, the known information changes and the agent is able to update the remaining features that it wishes to acquire (based on a new search with the addition information in $x_j$). Here we propose another heuristic  by ``breaking the tie'' in features in $u(x_o, o)$ according to the  single feature in $u(x_o, o)$ most minimizes the expected loss (\refalg{alg:aaco_alg}).
\begin{algorithm}[tb]
   \caption{Approximate Acquisition Conditioned Oracle Policy, with Tie-break Selection}
   \label{alg:aaco_alg}
\begin{algorithmic}
   \STATE {\bfseries Input:} Initial observed features $o$ (e.g., $o=\varnothing)$, instance values $x_o$, Data $\mathcal{D}=\{(x^{(i)}, y^{(i)})\}_{i=1}^n$, estimator $\hat{y}$.
   \vspace{4pt}
   \STATE Initialize \texttt{predict} $\coloneqq$ \texttt{false}
   \WHILE{$|o| < d$ {\bfseries and  not} \texttt{predict}}
   \STATE Draw $\mathcal{O} \subset \{v | v \subseteq \{1, \ldots, d\} \setminus o\}$ Uniformly
   \STATE $u(x_o, o) \coloneqq$ \resizebox{.7\hsize}{!}{$\argmin\limits_{v \in \mathcal{O}} \tfrac{1}{k} \sum\limits_{\scaleto{i \in N_k(x_o)}{5pt}} \ell\left( \hat{y}(x_{o}, \xind{i}_v), \ind{y}{i} \right) + \alpha c(o \cup v)$}
   \IF{$u(x_o, o) = \varnothing$}
   \STATE \texttt{predict} $\coloneqq$ \texttt{true}
   \ELSE
   \STATE $o = o\, \cup\, \left\{\argmin\limits_{w \in u(x_o, o)}  \sum\limits_{i \in N_k(x_o)} \ell\left( \hat{y}(x_o, \xind{i}_w), \ind{y}{i} \right)\right\}$
   \ENDIF
   \ENDWHILE
   \STATE {\bfseries Return} Prediction $\hat{y}(x_o)$
\end{algorithmic}
\end{algorithm}

\subsection{Behavioral Cloning}
The approximate ACO policy, $\hat\pi_\mathrm{ACO}(x_o, o)$, defined in \refalg{alg:aaco_alg} is a valid nonparametric policy in that, for a new instance drawn at test time, it is deployable and can actively acquire features and make a prediction without ever using unacquired features or the instance's label. 
One may, however, want a parametric policy $\pi_\theta(x_o, o)$ that is able to succinctly decide what actions to take without having to store and search through the training data (e.g. where $\pi_\theta$ is a parameterized function stemming from neural network weights $\theta$).
Fortunately, mimicking a  teacher policy is the focus of the well studied problem of imitation learning \citep{hussein2017imitation}, where we are able to leverage algorithms such as DAgger \citep{ross2011reduction} to supervise and train a parametric policy $\pi_\theta(x_o, o)$ based on the approximate ACO policy $\hat\pi_\mathrm{ACO}(x_o, o)$. 
In practice, we observed that a simple behavioral cloning approach \citep{bain1995framework} that directly trains $\pi_\theta(x_o, o)$ based on roll-outs of $\hat\pi_\mathrm{ACO}(x_o, o)$ was an effective way of supervising the parametric policy.
The surprising effectiveness of behavioral cloning is perhaps attributable to the ``non-cheating'' nature of the ACO, which uses only the same available information as $\pi_\theta(x_o, o)$ (and can thereby be easily emulated) (See \refsec{sec:exps}).

\subsection{AFA for Decision Making}

We now consider an important extension to settings
where one wishes to actively acquire features to determine a \emph{decision} that will maximize an \emph{outcome}, rather than for determining a \emph{prediction} to match a \emph{label}, as before.
For example, this is relevant in a medical context where clinicians need to decide between treatment strategies for a patient. A fundamental challenge in constructing algorithms for decision tasks is that the outcome under only one decision is ever observable for a given instance. Therefore, decision making requires constrasting counterfactuals that are unobserved in the data. The construction of policies to make optimal decisions from data with \emph{fully observed} contexts has been extensively studied in statistics, computer science, and operations research under the titles dynamic treatment regimes (DTRs) or individualized treatment rules (ITRs) \cite{murphydtrs, atheypolicy, kosorokpm}. 

Below, we develop analogies between components in a causal inference setup and ACO for prediction that enable AFA for general decision making using our framework.
Let $Y(a)$ denote the potential outcome \citep{rubin_po} under decision (intervention) $a\in \mathcal{A}$, where we assume larger values of $Y(a)$ are better without loss of generality. It can be shown that, under certain conditions (see Appendix), one can train a partially observed decision-making policy $\hat\pi_\mathcal{A}(x_o)$ which maps observed feature values $x_o$ to interventions to maximize $\mathbb{E}[Y(\hat\pi_\mathcal{A}(x_o))]$. Note that this decision-making policy $\hat\pi_\mathcal{A}(x_o)$ is analogous to $\hat{y}(x_o)$, the classifier given partially observed inputs, since $\hat\pi_\mathcal{A}(x_o)$ similarly maps partially observed inputs to outputs. Moreover, one can also train an estimator, $\hat{Q}(x,a)$, of $\mathbb{E}[Y(a) \mid x]$, the expected outcome for an instance $x$ with action $a$. 
$\hat{Q}(x,a)$ is akin to a loss function $\ell$ as it judges the effectiveness of an output $a$ for an instance.
Leveraging $\hat{Q}(x,a)$, and $\hat\pi_\mathcal{A}(x_o)$, it is now possible to utilize an approximate ACO to construct a decision-making acquisition policy $\pi_\mathrm{acq}(x_o)$, which determines what (if any) new features are worth acquiring in order to make a better decision (one that shall yield a higher expected outcome). Analogously to \eqref{eq:aco_search}, we may minimize:
\begin{equation}
    \resizebox{.9\hsize}{!}{$\argmin\limits_{v \subseteq \{1, \ldots, d\} \setminus o} \tfrac{1}{k} \sum\limits_{i \in N_k(x_o)} -\hat{Q}\left( \xind{i}, \hat{\pi}_\mathcal{A}\right(\xind{i}_{o \cup v}\left) \right) + \alpha c(o \cup v)$}, \label{eq:aco_dsearch}
\end{equation}
which uses $k$ neighbors $N_k(x_o)$ in the training set 
to determine if acquiring new features $v$ will lead to better decisions (made by $\hat{\pi}_\mathcal{A}$ and assessed using $\hat{Q}$). As before, the ACO's policy $\pi_{\tiny\mathrm{acq}}(x_o)$ 
 stemming from \eqref{eq:aco_dsearch} will determine what new features to sequentially acquire until reaching a subset $o^\prime$ for which no new acquisitions are worth the cost, at which time we choose an intervention according to $\hat\pi_\mathcal{A}(x_{o^\prime})$. To the best of our knowledge, this represents the first oracle based approach for AFA for general decision-making. 

\section{Experiments}
\label{sec:exps}

\begin{figure*}[ht]
    \centering
    \subfigure{\includegraphics[width=0.3\textwidth]{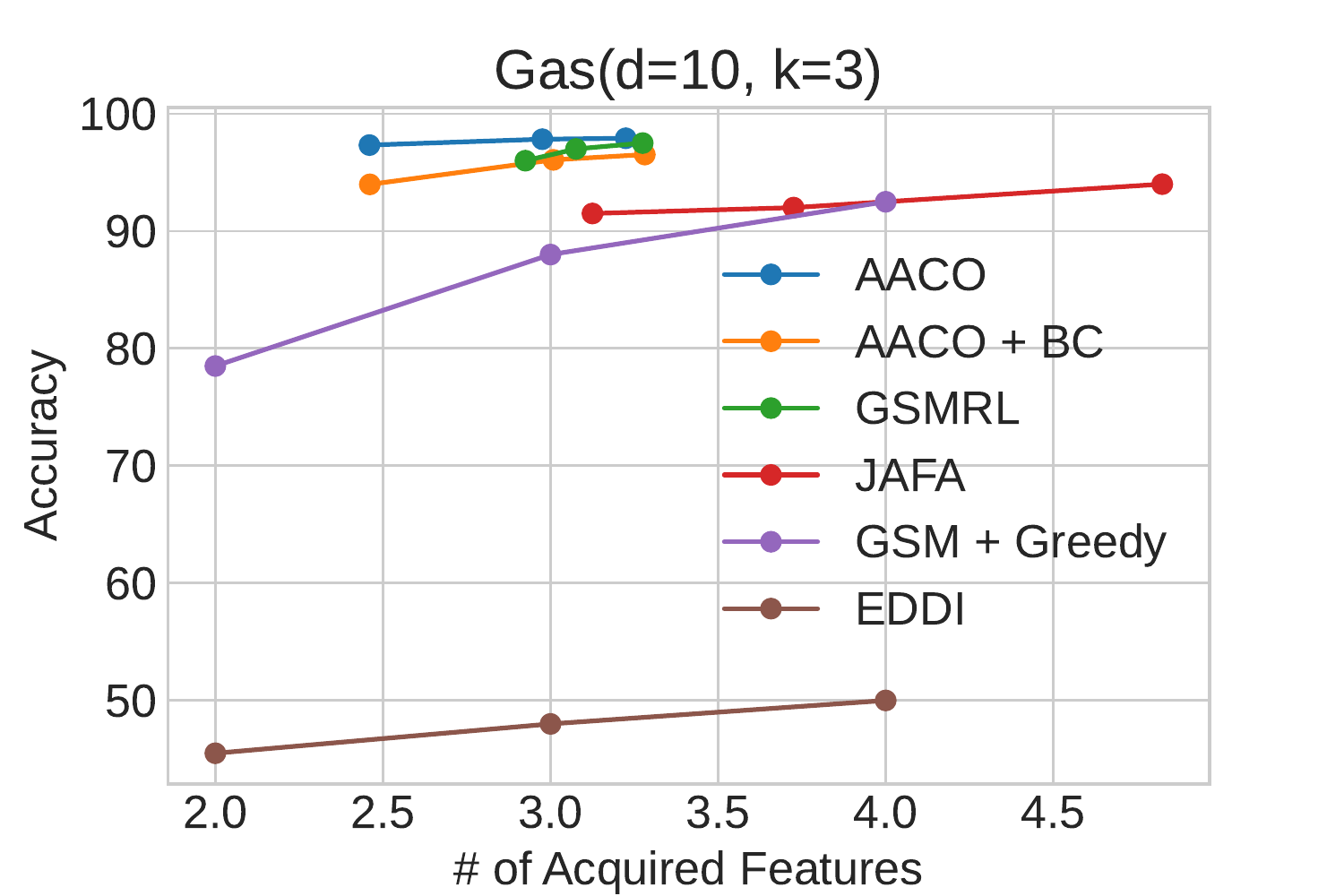}}
    \subfigure{\includegraphics[width=0.3\textwidth]{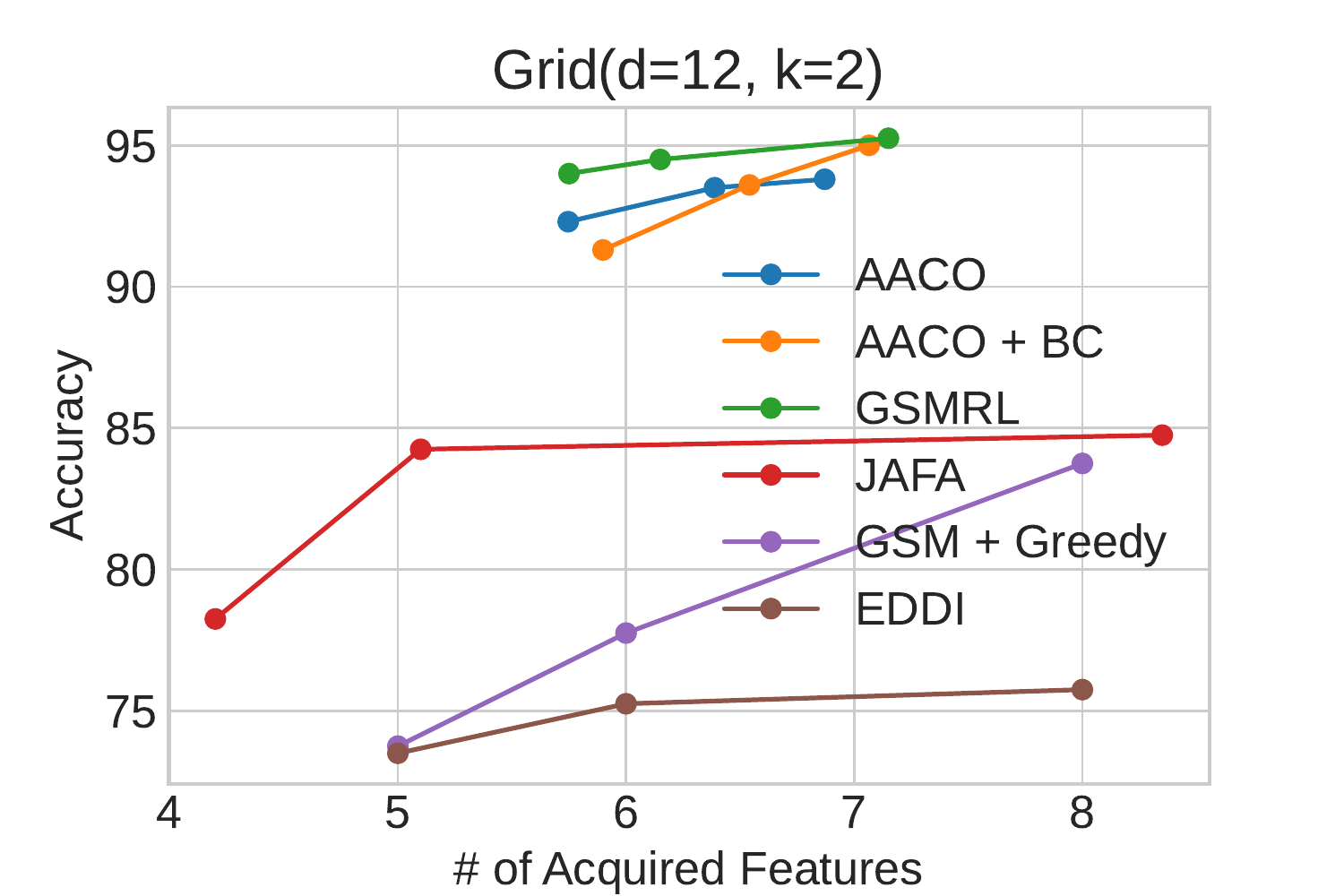}}
    \subfigure{\includegraphics[width=0.3\textwidth]{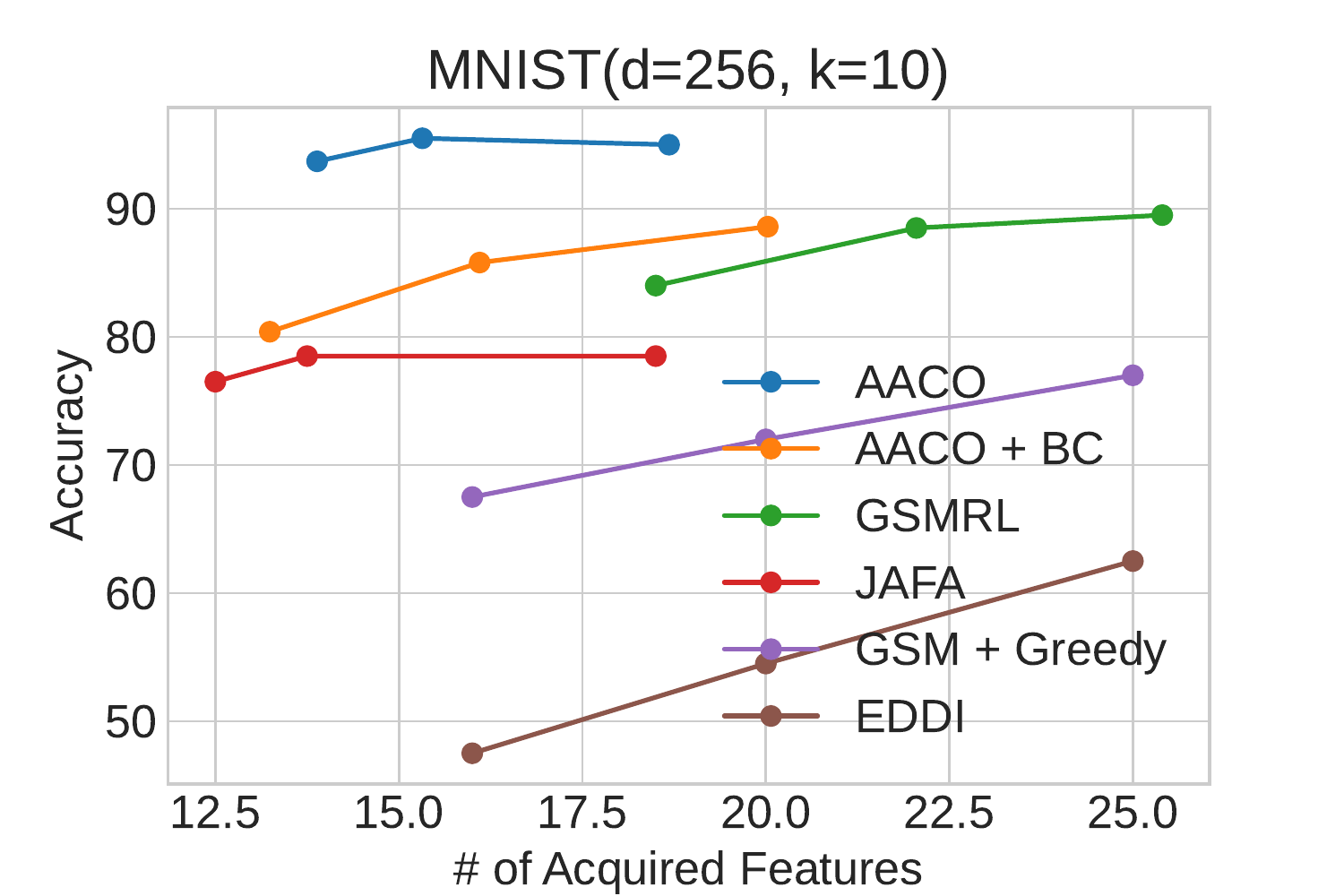}}
    \vspace{-8pt}
    \caption{Test accuracy on Real-world datasets.}
    \label{fig:cls_acc}
\end{figure*}


Next, we perform extensive experimentation to assess the ACO's performance relative to SOTA AFA methods.
Throughout, we compare to:
\texttt{JAFA} \citep{shim2018joint}, which jointly trains a deep learning RL agent (using Q-Learning) and a classifier for AFA;
\texttt{GSMRL} \citep{li2021active}, which learns a generative surrogate arbitrary conditioning model to derive auxiliary information and rewards that are used to train a deep learning agent; and other methods (described below). Please refer to \cite{shim2018joint, li2021active} for baseline hyperparameters details.
We use an equal cost for each feature and report multiple results with
different $\alpha$ \originaleqref{eq:aco_search} (as distinct ticks, e.g.~\reffig{fig:cubefeats}) to trade-off between task performance and acquisition cost. 
Code will be made open-sourced upon publication (see Supp. Mat.). 

We found that gradient boosted trees tended to perform well and were therefore used for the ACO policy (\texttt{AACO}, \refalg{alg:aaco_alg}, w/ cross-entropy $\ell$) predictor $\hat{y}$. The training procedure for $\hat{y}$ varied 
upon the dimensionality of the features. Small to moderate dimension settings allowed for the separate training of a $\hat{y}$ for each possible subset of features, leading to a dictionary of predictors. In higher dimensions, we utilized a masking strategy where the feature vector, whose unobserved entries were imputed with a fixed value, was concatenated with a binary mask that indicated the indices of observed entries \citep{li2020acflow}. During training, these binary masks were drawn at random to simulate making predictions with missing features. While this predictor allows for predictions with arbitrary subsets of features and could be used to make the final predictions, we found better performance by using the aforementioned dictionary-of-predictors approach on the subsets of features available at prediction, which was often significantly smaller than $2^d$ or $n$ due to the redundant subsets of features acquired.  The \texttt{AACO} policy approximates the distribution of $p(y, x_u | x_o)$ by using the set of $k$ nearest neighbors. In our experiments, we use $k =5$, with the exception of the Cube dataset, which exhibited better performance with a larger ($k=20$) number of neighbors. Further details are provided in Appendix.

\subsection{Synthetic Cube Dataset}

\vspace{0pt}
\begin{figure}[!ht]
    \centering
    \subfigure{\includegraphics[scale=0.21]{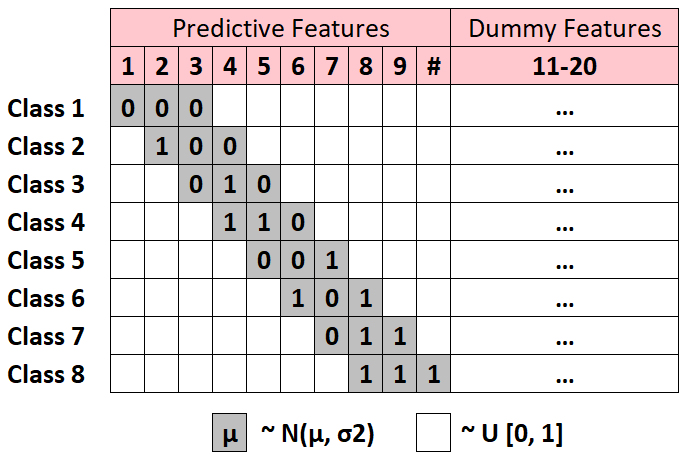}}
    \subfigure{\includegraphics[scale=0.27]{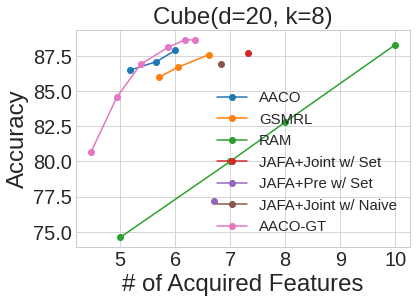}}
    \vspace{-10pt}
    \caption{Left: Distribution of features in CUBE-$\sigma$. Right: Accuracy for methods. Multiple ticks per same method represent different average number of acquired features vs. accuracy for policies stemming from different cost scales $\alpha$. }
    \vspace{-14pt}
    \label{fig:cubefeats}
\end{figure}

We begin with a study of the \texttt{CUBE-$\sigma$} dataset (as described by \citet{shim2018joint}),  a synthetic classification dataset designed for feature acquisition tasks. We consider a $d=20$-dimensional version with 8 classes. The feature values are a mixture of values drawn from one of 3 distributions: uniform $[0,1]$, $\mathcal{N}(0, \sigma^2)$, and $\mathcal{N}(1, \sigma^2)$; in our case, we chose $\sigma=0.3$ in order to match existing comparative baselines. Each data point consists of 17 uniform and 3 normally distributed features; the class label of the point determines the position and mean of the 3 normal features. For a data point of class $k$, features $k$ through $k+2$ will be normally distributed, with means as shown in \reffig{fig:cubefeats}; remaining features are uniform. 

We compare to previous approaches performing AFA on the \texttt{CUBE-$0.3$} dataset (as reported by \citet{shim2018joint}), which include: \texttt{JAFA+*} variations \citep{shim2018joint}; 
and \texttt{RAM} \citep{mnih2014ram}. The main Q-learning approach described by \citet{shim2018joint} trains all of its constituent networks in a joint fashion; the authors also consider ablations in which the jointly trained classifier is replaced with a pre-trained classifier and in which a naive feature encoding is used (keeping the structure of the classifier and value network consistent) instead of a learned set encoder. We also compare to \texttt{GSMRL}, based on their official repository implementation on \texttt{Cube} \citep{li2021active}. 
As can be seen from \reffig{fig:cubefeats}, despite not using deep RL methodology, AACO outperforms other SOTA approaches and nears the performance of an AACO policy with the ground truth classifier (\texttt{AACO-GT}), which is available in this synthetic dataset. 

\subsection{Real World Datasets}
\label{sec:real}

Next we perform experiments on real-world datasets stemming from the UCI ML repository and MNIST. 
In these experiments we additionally compare to:
\texttt{EDDI} \citep{ma2018eddi}, a greedy policy
that estimates the information gain for each candidate feature using
a VAE-based model with a set embedding and selects one feature with the highest expected utility at each acquisition step;
\texttt{GSM+Greedy} \citep{li2021active}, which similarly acquired features greedily using a surrogate arbitrary conditioning model that
estimates the utility of potential feature acquisitions.
Results of \texttt{JAFA}, \texttt{GSMRL}, \texttt{EDDI}, \texttt{GSM+Greedy} are reported from \cite{li2021active}.
Due to the higher dimensionality of MNIST, most baselines are unable to scale, and hence we consider a downsampled $16\times16$ version were $d=256$. (See below for an ablation on the full-dimensional MNIST.)

Here, unlike with the synthetic \texttt{Cube} dataset considered above, labels need not be predictable with only a few observed features. Surprisingly, it can be observed that one can make effective predictions using only a few features. 
Moreover, we see that our ACO approaches are often outperforming the state-of-the-art methods.
This is especially impressive considering that most of the baselines utilize complicated deep-learning approaches, where the \texttt{AACO} is utilizing simple nonparametric techniques.
Furthermore, note that the performance of the parametric policy \texttt{AACO+BC}, is competitive with that of the oracle \texttt{AACO}, despite being supervised using a relatively simple behavioral cloning approach.

\subsection{Ablations}
Next we perform some ablation experiments to study the effect of the oracle construction and higher dimensionalities.

\vspace{-0pt}
\begin{figure}[!h]
    \centering
    \subfigure{\includegraphics[width=0.48\linewidth]{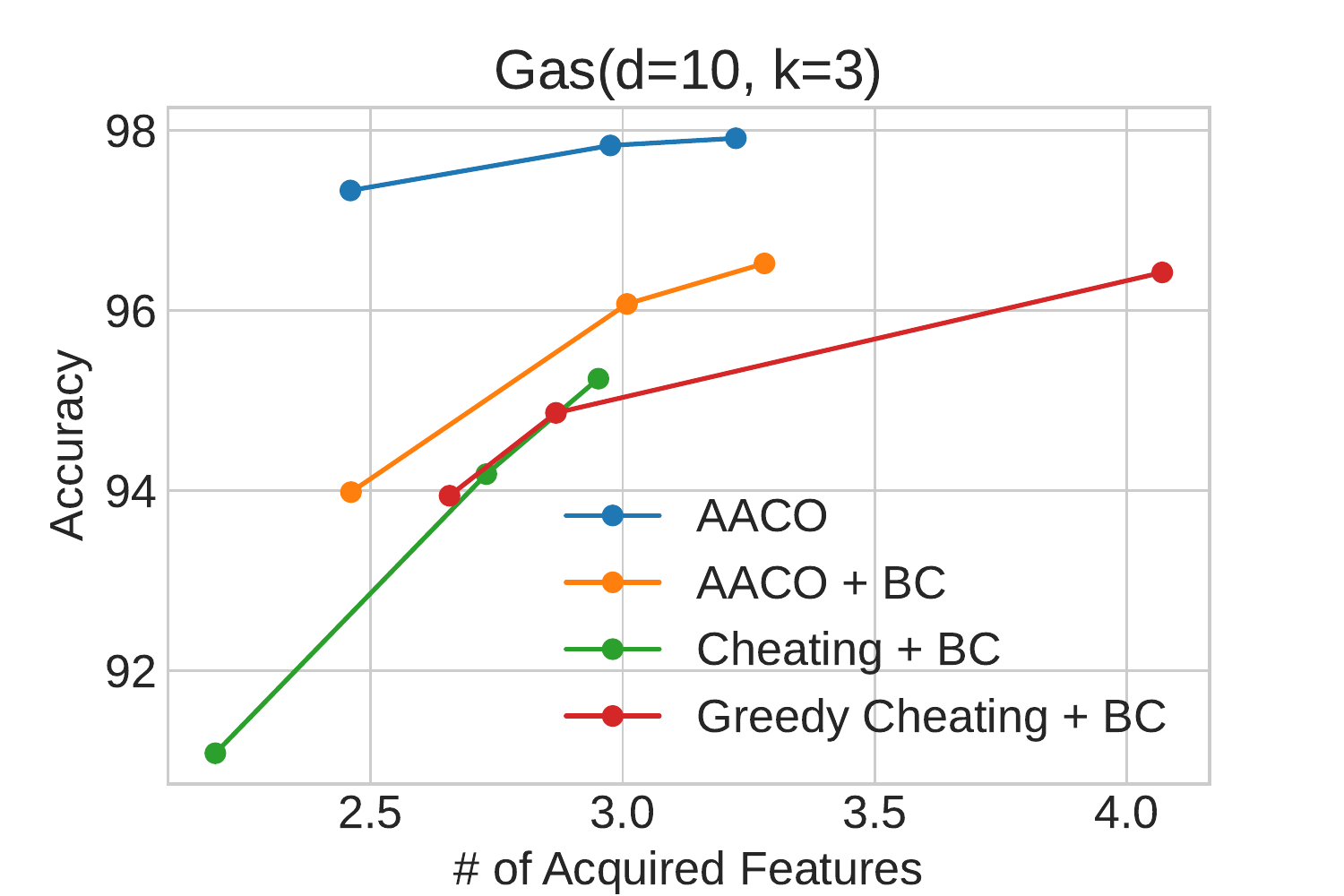}}
    \subfigure{\includegraphics[width=0.48\linewidth]{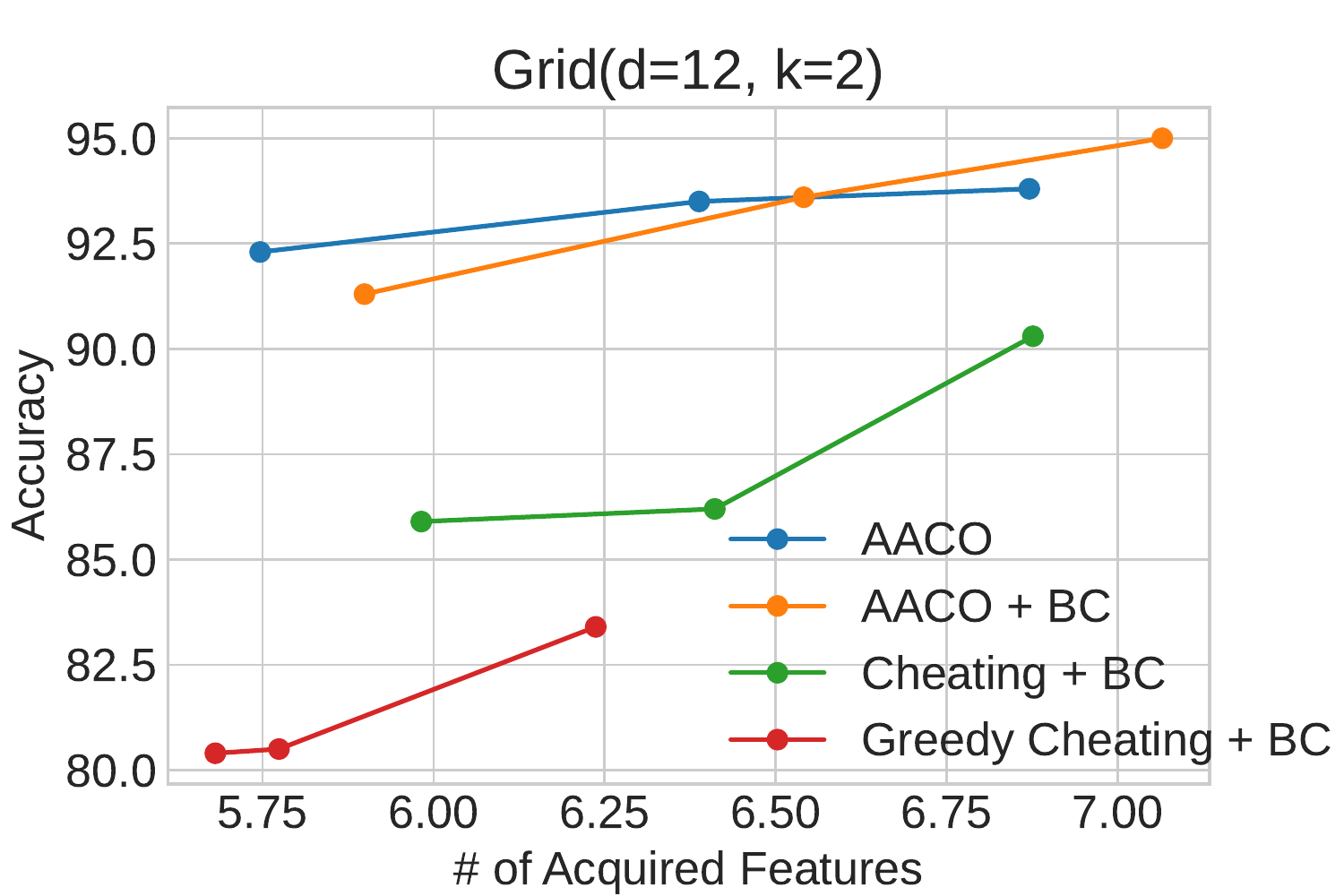}}
    \vspace{-10pt}
    \caption{Oracle Ablations.}
    \label{fig:oracle_ablations}
\end{figure}

\vspace{-0pt}
\textbf{Oracle Comparisons}\quad
As previously discussed, the cheating oracle, both in its greedy and non-greedy versions, operate on different state-spaces than an AFA policy since they observe information ($y$ and $x_u$) that is not accessible to an AFA agent. Thus, in order to utilize these cheating oracles, one must use them to supervise a parametric policy. Here we compare supervising the same (gradient boosted tree-based) parametric policy using behavioral cloning based on greedy-cheating \originaleqref{eq:greedy_search}, which is analogous to \cite{he2012imitation}, nongreedy-cheating \originaleqref{eq:nongreedy_search}, and AACO (\refalg{alg:aaco_alg}) oracles. As can be seen, behavior cloning consistently does a better job when mimicking the AACO policy. This is likely due not only to the non-greedy nature of ACO, but also due to fact that this teacher policy utilizes the same information as student and can thereby be more easily emulated.

\begin{wrapfigure}{r}{0.20\textwidth}
\vspace{-12pt}
    \centering
    \includegraphics[trim={5mm 0mm 5mm 0mm},clip,width=0.20\textwidth]{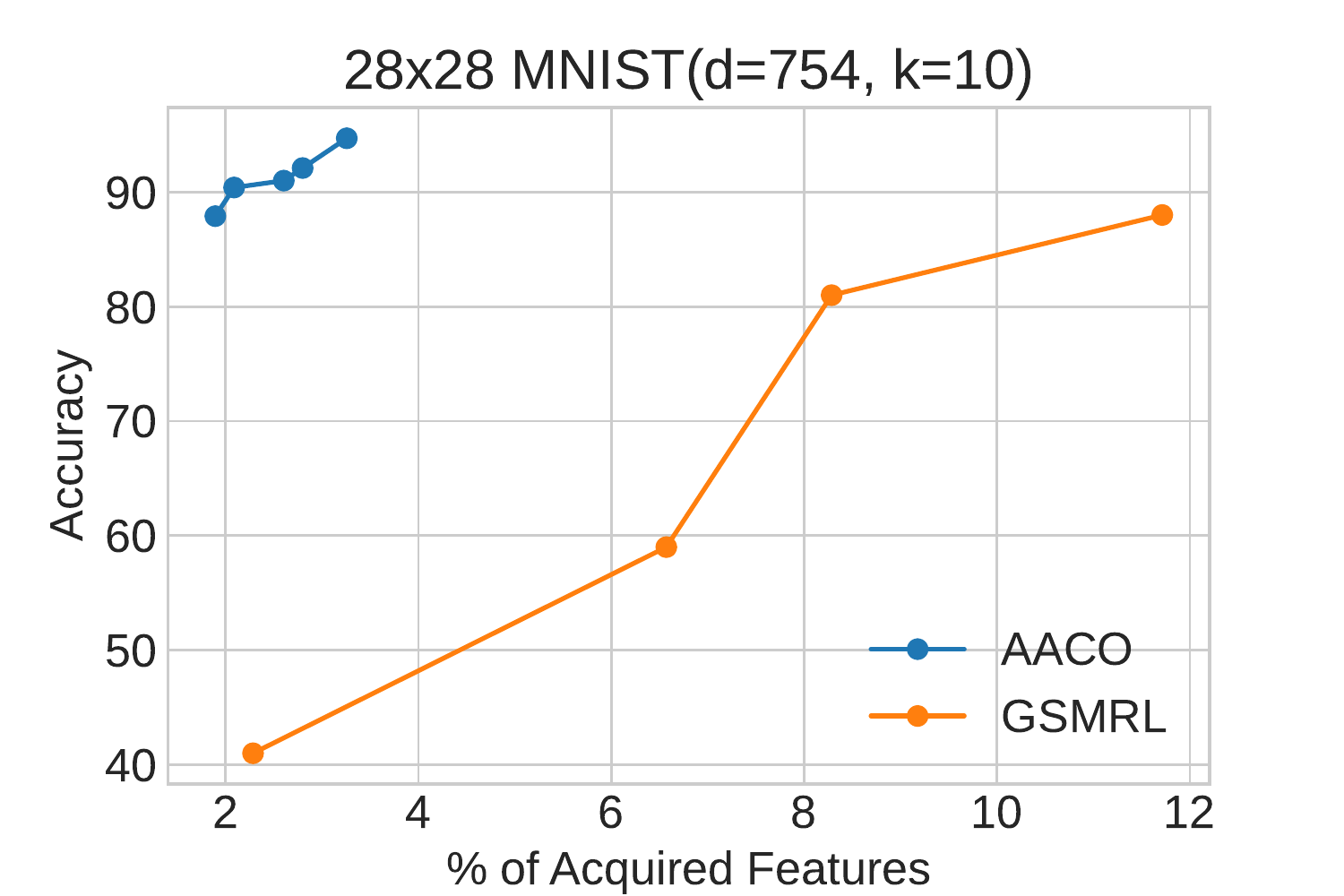}
    \vspace{-20pt}
    \caption{{\small $786d$} MNIST. }
    \label{fig:mnist_28}
\vspace{-12pt}
\end{wrapfigure}
\textbf{Higher Dimensionalities}\quad As noted by \citet{li2021active}, most existing AFA baseslines fail to scale to higher dimensional settings, such as the full $28 \times 28$ MNIST dataset. With 
$784$-d MNIST \texttt{JAFA} struggles to learn a policy that selects a small number of features (and instead learns to select all or no features) \citep{li2021active}. Furthermore, greedy methods like \texttt{EDDI} and \texttt{GSM+Greedy} are unable to scale their searches. Indeed, even \texttt{GSMRL}, which is able to learn a reasonable policy in the $784$-d MNIST, actually sees a degradation of performance when compared to the policy in $256$-d MNIST (\reffig{fig:cls_acc}). 
In contrast, 
we find that the AACO policy on the $784$-d MNIST achieves higher accuracy with significantly fewer features than \texttt{GSMRL} (Figure \ref{fig:mnist_28}). This finding, as well as the relative success on the $256$-d MNIST, provide evidence that the ACCO policy might better navigate high dimensional feature spaces than its competitors. 

\subsection{Decision Making}

\begin{wrapfigure}{r}{0.25\textwidth}
\vspace{-18pt}
    \centering
    \includegraphics[trim={5mm 0mm 5mm 0mm},clip,width=0.25\textwidth]{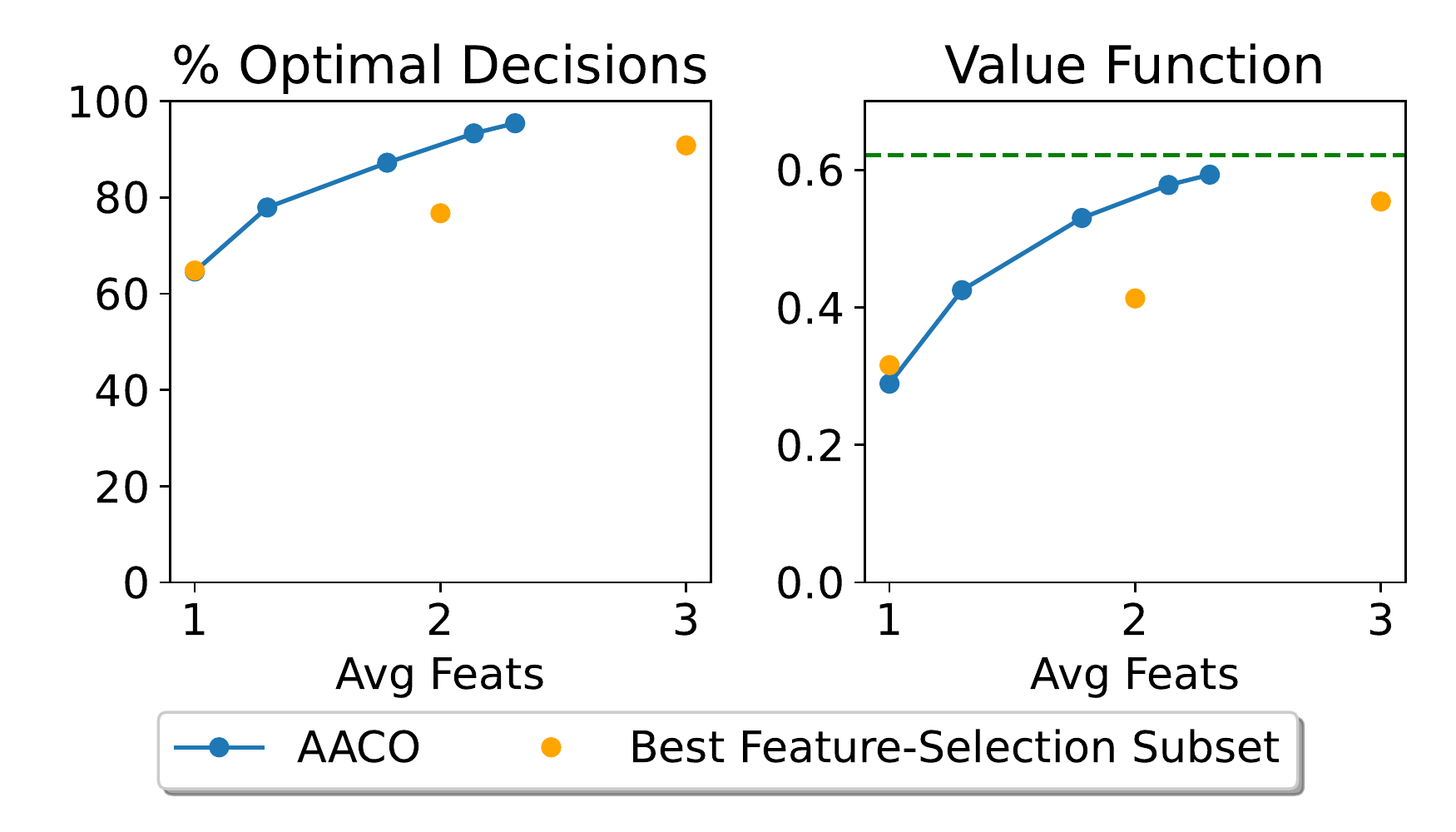}
    \vspace{-24pt}
    \caption{Decision making on synthetic data. Left: \% of time policy made the ground-truth optimal decision. Right: value function of the policy.}
    \label{fig:dec_syn}
\vspace{-10pt}
\end{wrapfigure}
We investigate the ability of the AACO's policy $\pi_{\mathrm{acq}}(x_o)$ to learn what features are relevant for making decisions that maximize expected outcomes according to a decision policy $\hat{\pi}_{\mathcal{A}}(x_{o})$. We first evaluate its performance under a synthetic environment, where the ground-truth optimal decisions are known (i.e. $Y(a)$ is known for every $a \in \mathcal{A}$). We generate four features $x_0,..,x_3$, a treatment $a \in \{0,1\}$, and an outcome $y$ that depends on $x_0,...,x_3$ and $a$. Full details of the environment are provided in the Appendix.  We use gradient boosted trees to fit $\hat{Q}(x,a)$ and separate decision policies $\hat{\pi}_{\mathcal{A}}(x_{o})$ for every $o$. As depicted in Figure  \ref{fig:dec_syn}, we find that increasing acquisition costs prompts the AACO policy to choose relevant features for decision making. By construction, an average of 2.25 features are sufficient for optimal decision making. At similar levels of AACO acquired features, the partially-observed decision-making policy ($\hat{\pi}_{\mathcal{A}}(x_o)$) attains near-perfect decision making with a value function ($\mathbb{E}[Y(\hat\pi_\mathcal{A}(x_o))]$) approaching the value function of the full-context optimal decision policy (green line). 
As AFA in this setting has been previously understudied, here we compare to a feature selection approach that uses a \emph{constant} subset of features (depicted in orange). This highlights AACO's ability to dynamically acquire relevant features on a per-instance basis, which leads to better performance.

\begin{wrapfigure}{r}{0.25\textwidth}
\vspace{-12pt}
    \centering
    \includegraphics[trim={2mm 0mm 5mm 0mm},clip,width=0.25\textwidth]{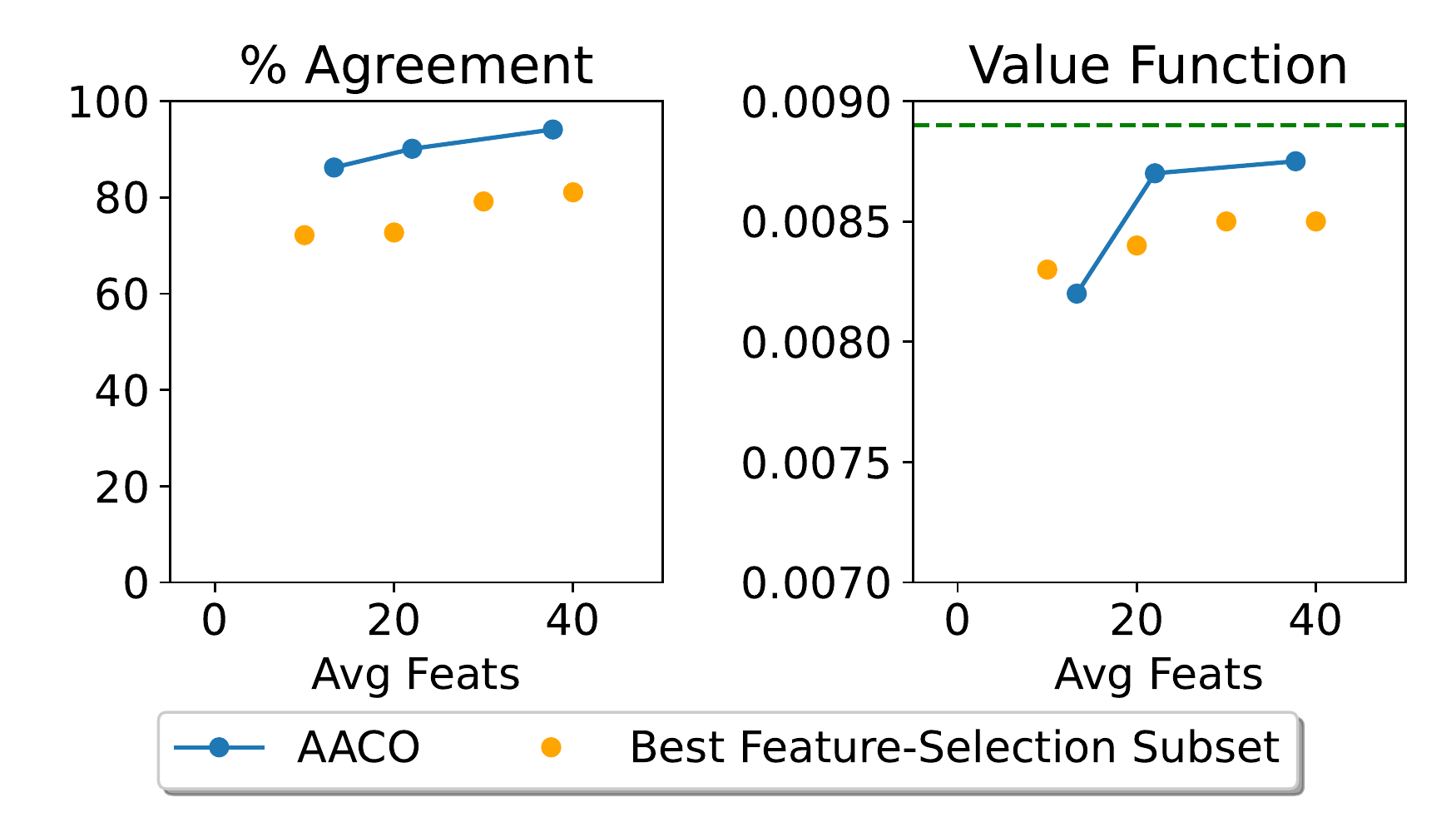}
    \vspace{-24pt}
    \caption{Left: \% agreement (in decision-making) with the full feature policy. Right: estimated value function. }
    \label{fig:dec_ovd}
\vspace{-10pt}
\end{wrapfigure}
Next, we demonstrate the practical utility of our method in the contextual bandit setting, which encompasses many decision making problems such as internet adverting \citep{cb_ads}, content recommendation \citep{cb_rec}, and medical treatment strategies \citep{kosorokpm}. When contexts are associated with a \textit{cost} (possibly in the form of time, computation, or money), a fruitful strategy is to acquire features based on their relevance for decision making. We illustrate a novel application of this end-goal 
using the Open Bandit Dataset \citep{saito2020open}. Here, we 
propose actively acquiring features detailing a user to recommend products (clothing items) in an attempt to maximize the expected click-rate by the user. The data set consists of over 1 million instances with 65 features. For simplicity, the action space is restricted to the two most frequent clothing item recommendations (see Appendix).

Here the ground-truth optimal decisions are unknown. Therefore, we measure the effectiveness of our methods relative to a learned recommendation policy based on full contexts. As a baseline, we compare our strategy to a natural cost-aware alternative, feature selection, in which a fixed subset of features are used by the decision-policy for all instances. The fixed subset of features is chosen to maximize the policy's estimated value ($n^{-1} \sum_{i=1}^n \hat{Q}(x^{(i)}, \pi_{\mathcal{A}}(x^{(i)}_o))$). In Figure \ref{fig:dec_ovd}, we see that the partially-observed decision policy under AACO makes the same decisions as the full-context policy over 90\% of the time with only 20 features, has an estimated value close to the full-context policy (green line), and generally outperforms the best feature selection alternative. This highlights the ability of the policy to make satisfactory decisions without observing full contexts and to tailor the acquired contexts to each instance.

\section{Discussion}
\label{sec:diss}

In conclusion, this work establishes several notable findings of interest.
First, we show that, using a relatively simple approach, it is possible to perform a non-greedy retrospective search for active feature acquisition.
Second, by utilizing the conditional data distribution, we show that it is possible to define an oracle that utilizes only the information that is available to the target AFA policy. 
Although previous work has considered AFA ``oracles,'' to the best of our knowledge the ACO is the first AFA oracle that utilizes only information available in the MDP state, making it the first oracle that is deployable at inference time. 
Third, we show that simple nonparametric techniques can be effectively used in place of the true data conditionals to deploy the ACO in practice. 
Fourth, we expound on how to utilize the ACO as a teacher to train a parametric policy.
Fifth, we generalize our ACO active feature acquisition approach to yield a policy that acquires useful features in order to make a useful decision (as opposed to prediction) based on a few observed features. 
Lastly, we conduct extensive empirical experiments showing the superiority of our approach to the state-of-the-art methodology for AFA. It is particularly notable that, despite utilizing simple, nonparametric techniques (which do not need deep learning based models) the ACO is able to outperform deep learning based approaches, even in higher dimensions.


\bibliography{refs}


\pagebreak

\section{Supplementary Material}

\subsection{Causal Inference and Decision Making}

Letting $Y(a)$ denote the potential outcome under decision (intervention) $a\in \mathcal{A}$, a decision policy $\pi_{\mathcal{A}}(x)$ is defined as a mapping from the features $x$ to an action $a \in \mathcal{A}$. We note that the policy, by definition, depends on which set of features $x$ are used to determine the action. For a class of decision policies $\Pi$, an optimal decision policy $\pi_{\mathcal{A}}^*(x)$ is any policy maximizing $\mathbb{E}[Y(\pi_{\mathcal{A}})]$, where $Y(\pi_{\mathcal{A}})$ is the potential outcome under the action recommended by $\pi_{\mathcal{A}}$. For some instances in the training data set, $Y(\pi_{\mathcal{A}})$ might be unobserved since the decision policy might recommend a different action than the one observed. In what follows, we will outline sufficient causal assumptions that allow counterfactual quantities such as $\mathbb{E}[Y(\pi_{\mathcal{A}})]$ to be expressed and estimated in terms of the observed data. 

We make the assumption that the complete features $x$ are sufficient to adjust for confounding. That is, we assume that either \textbf{(1)} interventions $a$ are marginally independent of $Y(a)$  or that \textbf{(2)} $Y(a) \perp a | x$. \textbf{(1)} is satisfied when $a$ is marginally randomized such as in randomized controlled trials or A/B tests while \textbf{(2)} occurs when $a$ depends on a behavior policy $\pi_b(x)$. Furthermore, we assume $p(a | x)$, the conditional distribution of interventions in the collected data, is greater than 0 for all $a \in \mathcal{A}$ and $x$ for which $p(x) > 0$. 

Under scenario $\textbf{(1)}$, we have that $\mathbb{E}[Y(a) | x_o, a] = \mathbb{E}[Y | x_o, a]$. Then, for a given context $x_o$, the optimal policy (when $\Pi$ is unrestricted), can be identified as
\begin{align*}
    \pi^*_{\mathcal{A}}(x_o) = \underset{a \in \mathcal{A}}{\text{argmax }} \, Q(x_o, a),
\end{align*}
 where $Q(x_o, a) \coloneqq E[Y | a, x_o]$. The identification of the optimal policy under scenario \textbf{(2)} is complicated because, in general, it is not true that $\mathbb{E}[Y(a) | x_o, a] = \mathbb{E}[Y | x_o, a]$. Instead, we have that $\mathbb{E}[Y(a) | x_o, a] = \mathbb{E}[ \mathbb{E}[Y | x, a] | x_o]$, which depends on the (unknown) distribution of $p(x | x_o)$. However, by contrasting $\mathbb{E}[Y(\pi^*_{\mathcal{A}})]$ with $\mathbb{E}[Y(\pi_{\mathcal{A}})]$ (i.e. examining the regret from using $\pi_{\mathcal{A}}(x_o)$ instead of $\pi^*_{\mathcal{A}}(x)$), we find that

\begin{align*}
    \pi^*_{\mathcal{A}}(x_o) = \underset{\pi_{\mathcal{A}} \in \Pi}{\text{argmin }} \, \mathbb{E}[Q(x, \pi_{\mathcal{A}}^*(x)) - Q(x, \pi_{\mathcal{A}}(x_o))],
\end{align*}

Therefore, a partial-information policy $\pi_{\mathcal{A}}(x_o)$ can be estimated by minimizing the empirical version of this loss, which takes the form of a weighted classification problem. Then, with this learned policy $\hat{\pi}_{\mathcal{A}}(x_o)$ that uses partially observed information, we can proceed with the approximate AACO as outlined in the Methods section. 

\subsection{Additional Experiment Details}

For the Cube and MNIST experiments, searching over all possible subsets $ v \subseteq \{1, \ldots, d\} \setminus o$ to find the best (additional) subset of features is infeasible. As discussed in the Methods section, we approximate this minimization by choosing random subsets of features $\mathcal{O} \subseteq \{v | v \subseteq \{1, \ldots, d\} \setminus o \}$. In our experiments, we chose $|\mathcal{O}| = 10,000$.

When getting nearest neighbors to perform the search over, we found that as few as $k=5$ neighbors performed well, and this is the number of neighbors used in the experiments. In general, increasing the number of neighbors can lead to neighbors having values $x_o$ less similar to the test instance's $x^{(i)}_o$, but having larger numbers of neighbors could lead to a better approximation of the conditional expectation of the loss, representing a bias-variance trade off. To standardize the relative importance of each feature, all features were mean-centered and scaled to have a variance of 1.

The AACO acquisition policy is a valid nonparametric policy in that, for a new instance drawn at test time, it is deployable and can actively acquire features and make a prediction without ever using unacquired features or the instance's label. In our experiments, we also considered using behavioral cloning \citep{bain1995framework} to train a parametric policy $\pi_\theta(x_o, o)$ that imitates the AACO policy. After rolling the AACO policy out on the validation dataset, we trained gradient boosted classification trees to mimic the actions in this data. Then, these trained classification-based policies were rolled out on a test data set. As with training the arbitrary classification models, we utilized a masking strategy where the feature vector, whose unobserved entries were imputed with a fixed value, was concatenated with a binary mask that indicated the indices of observed entries \citep{li2020acflow}.

The Open Bandit Dataset \citep{saito2020open} is a real-world logged bandit dataset provided by ZOZO, Inc., a Japanese fashion e-commerce company. The data contains information collected from experiments where users are recommended one of 34 fashion items, with the response variable being whether or not the user clicked on the recommended item. The Open Bandit Dataset contains several campaigns under different recommendation policies. We analyzed the AACO policy under the Men's campaign with the Thompson Sampling policy. To simplify the presentation, we filtered the data to only include events where the two most frequently recommended products were recommended. This leads to a setting where the objective is to learn a binary decision policy that chooses between these two clothing recommendations. Altogether, the filtered data set has over 1 million instances and 65 features (i.e. the context $x$).

\subsection{Synthetic Decision-Making Environment}

In the synthetic decision-making environment, we create 4 features $(x_0, x_1, x_2, x_3)$, where $x_0 \sim U(0,1)$, $x_1$ follows a Rademacher(0.5) distribution, and $(x_3, x_4)$ are jointly normal with a correlation of 0.3. Furthermore, to imitate the realistic scenario in which interventions $a$ in a previously collected data set are not randomized, we draw $a$ a a Bernoulli random variable according to a probit model with a linear dependence on the four features. To create a setup where different numbers of features are relevant to decision making, we draw the outcome $y$ from a normal distribution with a mean equal to the following:

\begin{align*}
    a*\big[&\mathrm{I}(0 < x_0 \leq 0.25) + \mathrm{I}(0.25 < x_0 \leq 0.5) x_1 \\
    &+ \mathrm{I}(0.5 < x_0 < 0.75) x_1 x_2 \\
    &+ \mathrm{I}(0.75 < x_0 < 1) x_1 (x_3^2 - 1) \big]
\end{align*}

Clearly, the optimal decision when all of $x_1$, $x_2$, $x_3$, and $x_4$ are observed is to assign $a=1$ when the above quantity in brackets is positive and assign $a=0$ otherwise. The optimal decision when only a proper subset of the four features are observed requires marginalizing the bracketed quantity over the unknown features. Furthermore, the number of features relevant for decision making varies from instance to instance. For example, if $0 < x_0 \leq 0.25$, then only $x_0$ is needed to make an optimal decision. The feature $x_0$ is particularly important, since it is informative about which other features should be acquired.

\subsection{AACO Feature Acquisitions on CUBE}

\begin{figure}[!t]
\includegraphics[scale=0.25]{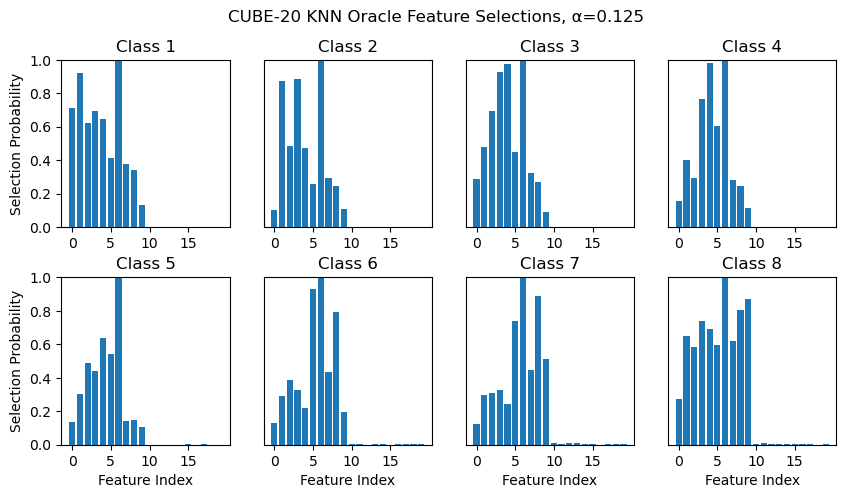}
\caption{Acquisition actions taken by AACO, expressed as a histogram per class for each feature. I.e., we report the portion of the time that a feature was acquired on average for making a prediction for instances of each class. Note that, in addition to not acquiring the noise features 10-19 in most cases, the agent tends to focus on the most defining features for each specific class, as indicated by Figure 3 in the main paper. (Here feature 6 was deterministically chosen as the initial feature.)}
\label{fig:cubeacq}
\end{figure}

We show the features acquired on the CUBE synthetic data (Fig.~\ref{fig:cubeacq}). We see that the features that were acquired follow the most defining features for each class.

\end{document}